\documentclass[times, review, 10pt]{elsarticle}

\usepackage{amssymb}
\usepackage{amsmath}
\usepackage{algorithmic}
\usepackage{algorithm}
\usepackage{subcaption} 
\usepackage{changepage}
\usepackage{bm}
\usepackage{booktabs}
\usepackage{multirow}
\usepackage{tabularx}
\usepackage{hyperref}
\usepackage{xcolor}
\definecolor{darkgray}{gray}{0.4}
\journal{Pattern Recognition}

\begin{document}

\begin{frontmatter}

\title{Deep Neural Network Calibration by Reducing Classifier Shift with Stochastic Masking}

\author[addr1]{Jiani Ni}
\ead{nijn23@mails.jlu.edu.cn}

\author[addr2]{He Zhao}
\ead{he.zhao@data61.csiro.au}

\author[addr3]{Yibo Yang}
\ead{yibo.yang93@gmail.com}

\author[addr1]{Dandan Guo\corref{cor1}}
\ead{guodandan@jlu.edu.cn}

\cortext[cor1]{Corresponding author}

\affiliation[addr1]{organization={School of Artificial Intelligence, Jilin University},
            city={Changchun},
            country={China}}

\affiliation[addr2]{organization={CSIRO's Data61 and Monash University (adjunct)},
            country={Australia}}

\affiliation[addr3]{organization={King Abdullah University of Science and Technology},
            country={Saudi Arabia}}

\tnotetext[grant]{This work of Dandan Guo was supported by the National Natural Science Foundation of China (NSFC) under Grant 62306125.}

%% Abstract
\begin{abstract}
In recent years, deep neural networks (DNNs) have shown competitive results in many fields.  
Despite this success, they often suffer from poor calibration, especially in safety-critical scenarios such as autonomous driving and healthcare, where unreliable confidence estimates can lead to serious consequences.  
Recent studies have focused on improving calibration by modifying the classifier, yet such efforts remain limited.  
Moreover, most existing approaches overlook calibration errors caused by underconfidence, which can be equally detrimental.
To address these challenges, we propose \textbf{MaC-Cal}, a novel mask-based classifier calibration method that leverages stochastic sparsity to enhance the alignment between confidence and accuracy.  
MaC-Cal adopts a two-stage training scheme with adaptive sparsity, dynamically adjusting mask retention rates based on the deviation between confidence and accuracy.  
Extensive experiments show that MaC-Cal achieves superior calibration performance and robustness under data corruption, offering a practical and effective solution for reliable confidence estimation in DNNs.
\end{abstract}

%%Research highlights
% \begin{highlights}
% \item A masking-based method effectively improves model calibration by reducing classifier shift.
% \item Two-stage training strategy clearly decouples feature learning from confidence calibration.
% \item Adaptive sparsity adjusts masks by matching confidence to accuracy.
% \item The method effectively mitigates both over- and underconfidence issues.
% \item Lightweight design ensures easy integration across models and datasets.
% \end{highlights}


\end{frontmatter}

\section{Introduction}
\label{sec:intro}

In recent years, deep neural networks (DNNs) have achieved remarkable progress in computer vision, with widespread applications in image classification \cite{LI2025111726,LI2025111822}, object detection \cite{LI2026112118,YANG2026112127}, and natural language processing \cite{vaswani2017attention,devlin2018bert}. However, beyond accuracy, the reliability of model confidence is equally critical, especially in high-stakes scenarios such as autonomous driving \cite{GUO2025111174,MADHAVI2025112154}, medical diagnosis \cite{HU2025111619,JIANG2026112152}, and credit risk prediction \cite{XU2021108125}.
Ideally, model confidence should align with the actual likelihood of correctness. For instance, predictions with a confidence of 0.7 should be correct approximately 70\% of the time. In practice, however, many DNNs suffer from calibration errors—predicted confidences often fail to match true accuracies and tend to be overconfident \cite{guo2017calibration}—which impairs downstream decision-making.

To address the mismatch between predicted confidence and true accuracy, calibration methods have attracted increasing attention. These methods are typically categorized into training-time and post-hoc approaches. Among the latter, Temperature Scaling \cite{guo2017calibration} adjusts the softmax temperature after training. While it performs well under in-distribution settings, it lacks robustness when facing distribution shifts \cite{ovadia2019can}. In contrast, training-time methods incorporate calibration directly into the model learning process. For instance, focal loss \cite{lin2017focal}, label smoothing \cite{szegedy2016rethinking}, and Mixup \cite{zhang2018mixup}—originally developed to improve generalization—have demonstrated secondary benefits in enhancing calibration. More recent approaches \cite{muller2019when,thulasidasan2019mixup,ghosh2022adafocal,cheng2022calibrating,liu2022devil,pinto2022using,tao2023calibrating,tao2023dual} further advance this direction by introducing explicit regularization strategies that directly constrain model confidence or uncertainty estimates.

Existing studies show that classifier structure significantly affects model calibration. 
One line of work improves calibration by decoupling feature extraction from classification \cite{jordahn2024decoupling}, while another adopts progressive layer freezing based on the weak classifier hypothesis \cite{wang2024calibration}. 
Though effective, both of them introduce trade-offs between flexibility, generalization, and calibration, and often show inconsistent results across architectures and datasets. 
Dual-head designs, such as fixed Simplex Equiangular Tight Frame (ETF) heads \cite{wang2024calibration}, further improve calibration by introducing an auxiliary classifier. Moreover, most methods mainly address overconfidence, overlooking calibration errors due to underconfidence.

Building upon these findings, we further investigate the classifier's role in model calibration. Our statistical analysis of uncalibrated model outputs shows a consistent confidence deviation from true accuracy. Further experiments suggest this calibration shift primarily originates from the classifier itself. We show that lightweight structural modifications to the classifier can effectively reduce such deviation. By mitigating classifier shift, both overconfidence and underconfidence may be alleviated. 
Motivated by limitations of existing approaches, we aim to design a lightweight, classifier-specific training method that improves calibration without disrupting feature learning.

To this end, we propose \textbf{MaC-Cal} (Mask-based Classifier Calibration), a calibration framework that enhances confidence reliability by introducing stochastic sparsity into the classifier. MaC-Cal employs a two-stage training process: the first stage jointly trains the feature extractor and classifier, while the second stage freezes the extractor and retrains the classifier under a masking.
To enhance flexibility and avoid the instability of fixed sparsity, we further introduce an adaptive sparsity mechanism, where the mask retention probability is dynamically adjusted based on the deviation between predicted confidence and actual accuracy. Our method is simple but effective, model-agnostic, and easy to be integrated into existing pipelines. It works across different architectures and datasets. The main contributions of this work are summarized as follows:

\begin{itemize}
    \item We explore the classifier’s role in model calibration and introduce a new   solution from the view of  classifier shift reduction.
    % stochastic sparsity, applied via masking.
    \item We propose a two-stage training framework with an adaptive masking mechanism to improve both over- and underconfidence in DNNs.
    \item Extensive experiments demonstrate that ours achieves superior performance while achieving robustness in various settings.
\end{itemize}

\section{Related work}
\label{sec:related}

\subsection{Confidence Calibration}

Calibration aims to mitigate the mismatch between predicted confidence and true accuracy, which is often manifested as overconfidence in deep neural networks. Calibration methods can be broadly categorized into two types: training-time and post-hoc approaches. 
Temperature scaling (TS) \cite{guo2017calibration} is a widely used post-hoc calibration method that adjusts the softmax temperature parameter after training. Although TS performs well under in-distribution settings, \cite{ovadia2019can} show that it exhibits limitations when applied under distribution shifts.
In contrast, training-time approaches incorporate calibration into the learning process. Techniques such as focal loss (FL) \cite{lin2017focal}, label smoothing (LS) \cite{szegedy2016rethinking} and Mixup \cite{zhang2018mixup} were originally proposed to enhance generalization and have been found to also improve calibration. Several recent training-time methods \cite{ni2025balancing,tao2023calibrating,tao2023dual,ghosh2022adafocal,cheng2022calibrating,liu2022devil,pinto2022using} explicitly introduce regularization on predicted probabilities to promote calibrated confidence estimates. 
In this work, we focus on training-time calibration approaches.

Recent training-time methods have increasingly addressed calibration by modifying the classifier itself. One representative approach decouples feature extraction from classification \cite{jordahn2024decoupling}, achieving promising calibration on in-distribution (ID) data, but suffering from poor generalization to out-of-distribution (OOD) inputs. Another study \cite{wang2024calibration} reveals that some training-time strategies can hinder a model’s ability to be calibrated post hoc. To mitigate this, a weak classifier hypothesis and the Progressive Layer-Peeled (PLP) training paradigm were proposed, incrementally freezing upper layers to support better post-hoc calibration. However, PLP's effectiveness is highly sensitive to model architecture and dataset, and may even impede convergence. More recent work \cite{ni2025balancing} integrates a fixed Simplex Equiangular Tight Frame (ETF) classifier with a standard linear classifier, forming a dual-head architecture that jointly balances their outputs for improved calibration. Yet, the method’s dependence on architectural modifications limits its real-world applicability. Building on these insights, we propose a classifier-driven calibration framework that minimizes structural changes while ensuring strong calibration and generalization on both ID and OOD data.

Moreover, while calibration errors are often caused by overconfidence, underconfidence is also a non-negligible factor, though less common. Mixup has been shown to improve both accuracy and calibration, but its calibration performance is highly sensitive to the interpolation strength—high mixing ratios can lead to underconfidence \cite{thulasidasan2019mixup}. Although \cite{wang2023pitfall} partially addresses this issue, their method is limited to Mixup and does not generalize well to other calibration techniques. In contrast, our method alleviates both overconfidence and underconfidence, providing a more robust and broadly applicable calibration solution across varied training setups.

\subsection{Dropout-based Methods}

Dropout, originally proposed by \cite{hinton2012improving}, is a widely used regularization technique that mitigates overfitting in neural networks by randomly dropping a subset of neurons during training. Over the years, numerous variants have been developed to better suit different architectures and training paradigms. These include DropConnect, which randomly removes connections between neurons \cite{wan2013regularization}, and input dropout mechanisms such as DropoutNet \cite{volkovs2017dropoutnet}. 
For convolutional architectures, spatially structured variants like DropBlock \cite{tompson2015efficient} and SpatialDropout \cite{ghiasi2018dropblock} have been proposed, while recurrent and graph neural networks have inspired structured dropout techniques \cite{moon2015rnndrop,semeniuta2016recurrent}. In deep residual networks, methods such as Stochastic Depth \cite{huang2016deep} and ShakeDrop \cite{yamada2019shakedrop} are tailored to improve training robustness.
Although these methods enhance generalization and offer limited improvements in uncertainty estimation, studies have shown that the resulting epistemic uncertainty remains poorly calibrated and performs unreliably in detecting OOD inputs \cite{gal2017concrete,lakshminarayanan2017simple}.

Recent works have explored dropout in calibration. \cite{zhang2019confidence} compared calibration errors across various dropout techniques, finding their effectiveness strongly depends on dropout rate, network architecture, and depth. This sensitivity limits their generalization ability.
\cite{laves2020calibration} proposed a post-hoc method to correct dropout-induced calibration errors but with limited generality. 
% To address these issues, leveraging prior insights on classifier-calibration relationships, we propose a general dropout-like calibration strategy using two-stage optimization directly on classifier weights, reducing dependence on network structure. An adaptive sparsity mechanism replaces fixed dropout rates, avoiding manual tuning and instability, achieving superior calibration without extra post-processing.
To address these issues, we propose a dropout-inspired calibration method that is inherently insensitive to architectural variations. The introduced adaptive sparsity significantly reduces the need for manual tuning.
\section{Background}
\label{sec:back}

% Consider a dataset \( D = \{(x_i, y_i)\}_{i=1}^{N} \), where \( x_i \in \mathcal{X} \) is an input sample and \( y_i \in \mathcal{Y} \) is its corresponding label from \( K \) classes. Let \( N \) be the total number of samples, and \( n_k \) be the number of samples in class \( k \), such that \( N = \sum_{k=1}^{K} n_k \).
% We model the deep learning system as comprising a feature extractor \( f: \mathcal{X} \to \mathbb{R}^d \) with parameters \( \bm{\theta} \), and a linear classifier \( \mathbf{W} = \{\bm{w}_k\}_{k=1}^{K} \in \mathbb{R}^{d \times K} \). The feature extractor maps the input \( x_i \) to a \( d \)-dimensional feature vector \( \bm{z}_i = f(x_i; \bm{\theta}) \), referred to as the last-layer feature. The classifier then produces the logit vector:
% \begin{equation}
% \bm{l}_i = \bm{z}_i \mathbf{W} \in \mathbb{R}^K,
% \end{equation}
% from which the predicted label \( \hat{y}_i \) and associated confidence score \( \hat{p}_i \) are derived.

Consider a dataset \( D = \{(x_i, y_i)\}_{i=1}^{N} \), where \( x_i \in \mathcal{X} \) is an input sample and \( y_i \in \mathcal{Y} \) is its corresponding label from \( K \) classes. Let \( N \) denote the total number of samples, and \( n_k \) be the number of samples in class \( k \), such that \( N = \sum_{k=1}^{K} n_k \).
We use a feature extractor \( f: \mathcal{X} \to \mathbb{R}^d \) with parameters \( \bm{\theta} \), and a linear classifier \( \mathbf{W} = \{\bm{w}_k\}_{k=1}^{K} \in \mathbb{R}^{d \times K} \) as the model components.
The feature extractor maps an input \( x_i \) to a \( d \)-dimensional representation \( \bm{z}_i = f(x_i; \bm{\theta}) \), denoted as the last-layer feature. The classifier then computes the logit:
\begin{equation}
\bm{l}_i = \bm{z}_i \mathbf{W} \in \mathbb{R}^K.
\end{equation}
The softmax function is then applied to obtain the predicted probability:
\begin{equation}
\bm{p}_i = \mathrm{softmax}(\bm{l}_i), \quad \text{where} \quad p_{i,k} = \frac{e^{l_{i,k}}}{\sum_{j=1}^{K} e^{l_{i,j}}}.
\end{equation}
Based on these probabilities, the predicted label \( \hat{y}_i \) and confidence \( \hat{p}_i \) are:
\begin{equation}
\hat{y}_i = \arg\max_k p_{i,k}, \quad \hat{p}_i = \max_k p_{i,k}.
\end{equation}

The model is typically trained using cross-entropy loss, which has been shown to yield calibration issues such as overconfidence \cite{guo2017calibration}. In addition, mixup training tends to produce underconfident predictions due to the interpolation between samples \cite{thulasidasan2019mixup}.

Calibration methods aim to align predicted confidence with true accuracy. Specifically, a model is perfectly calibrated if for all \( p \in [0,1] \), \( P(\hat{y} = y \mid \hat{p} = p) = p \)  \cite{guo2017calibration}.  
To quantify this alignment, the Expected Calibration Error (ECE) measures the expected difference between accuracy and confidence:

\begin{equation}
\mathrm{ECE} = \mathbb{E}_{\hat{p}} \left[ \left| P(\hat{y} = y \mid \hat{p}) - \hat{p} \right| \right].
\end{equation}
Since the true conditional probabilities are not directly observable, ECE is approximated via confidence-based binning: predictions are grouped into \( M \) bins \( \{B_m\}_{m=1}^M \) based on confidence intervals \(\left[ \frac{m-1}{M}, \frac{m}{M} \right) \), with average confidence \( C_m \) and accuracy \( A_m \) calculated for each bin. ECE is then computed as the weighted absolute difference between them:

\begin{equation}
\mathrm{ECE} = \sum_{m=1}^M \frac{|B_m|}{N} \left| A_m - C_m \right|.
\end{equation}
Besides the standard ECE, variants have been proposed to improve estimation stability. For example, Adaptive-ECE (AECE) \cite{nguyen2015posterior} groups samples into bins with equal sample counts rather than equal confidence intervals, and Maximum Calibration Error (MCE) \cite{guo2017calibration} measures the largest calibration deviation, focusing on the worst-case calibration performance.
% Beyond the standard ECE, several variants have been proposed to improve the robustness of calibration assessment. One notable extension is Adaptive Expected Calibration Error (AECE) \cite{nguyen2015posterior}, which modifies the binning approach by grouping samples into bins containing an equal number of data points rather than using fixed confidence intervals. This adaptive binning scheme aims to stabilize the estimation, especially when confidence distributions are skewed or uneven. Another important metric is the Maximum Calibration Error (MCE) \cite{guo2017calibration}, which instead focuses on the worst-case calibration error by measuring the maximum absolute difference between accuracy and confidence over all bins. 
\section{Methods}
\label{sec:methods}

\begin{figure}[t]
  \centering
  \subfloat[\label{fig:1a}]{\includegraphics[width=0.33\textwidth]{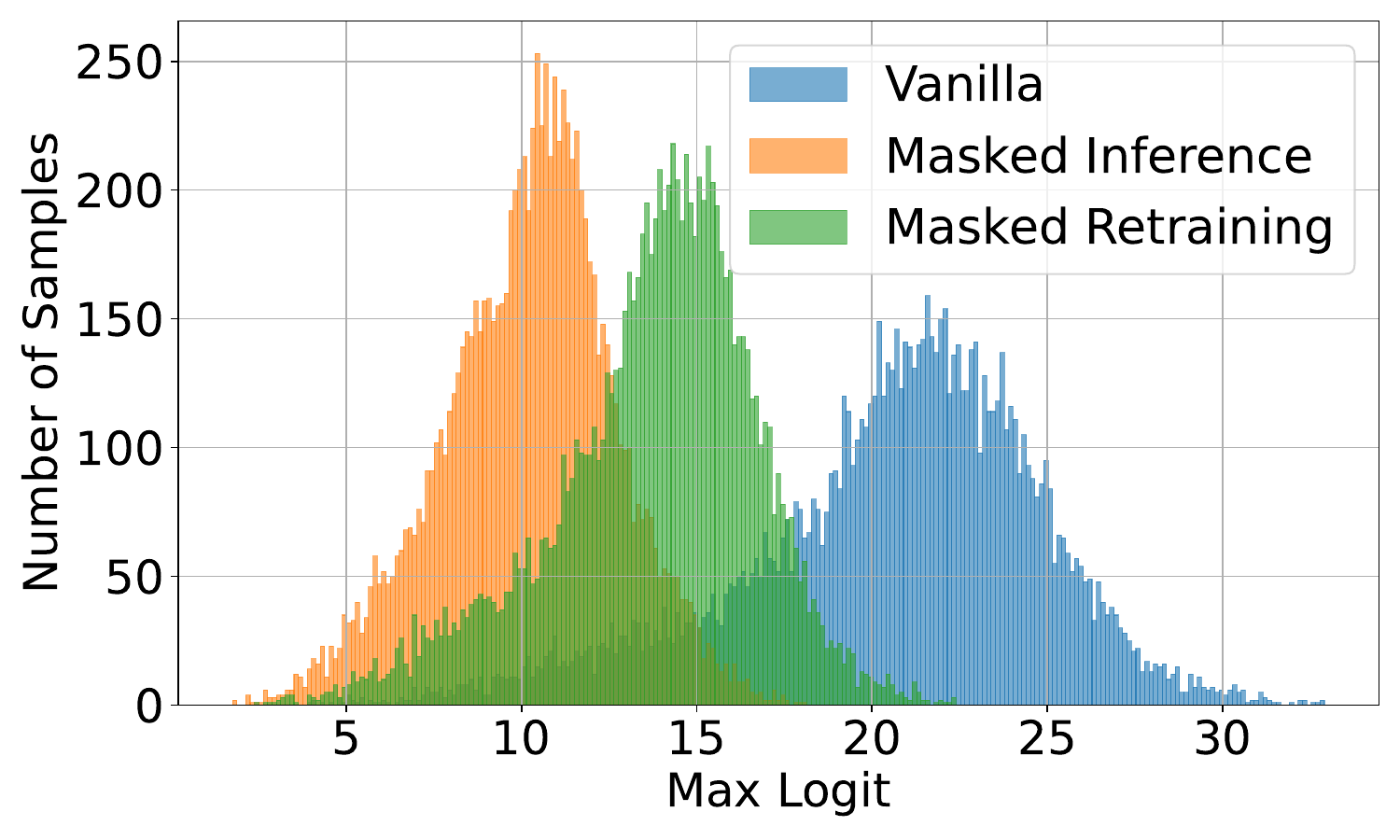}}
  \hfill
  \subfloat[\label{fig:1b}]
  {\includegraphics[width=0.33\textwidth]{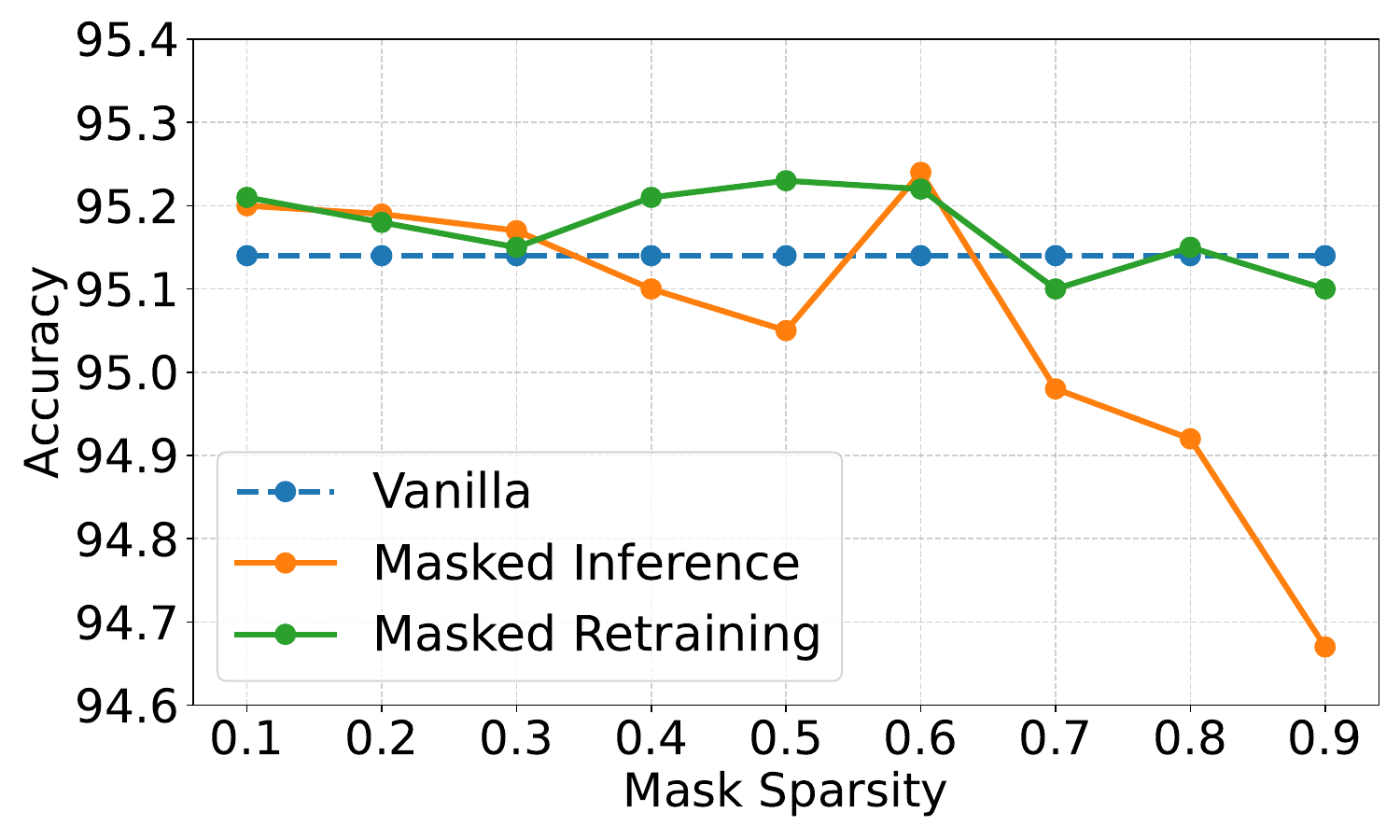}}
  \hfill
  \subfloat[\label{fig:1c}]
  {\includegraphics[width=0.33\textwidth]{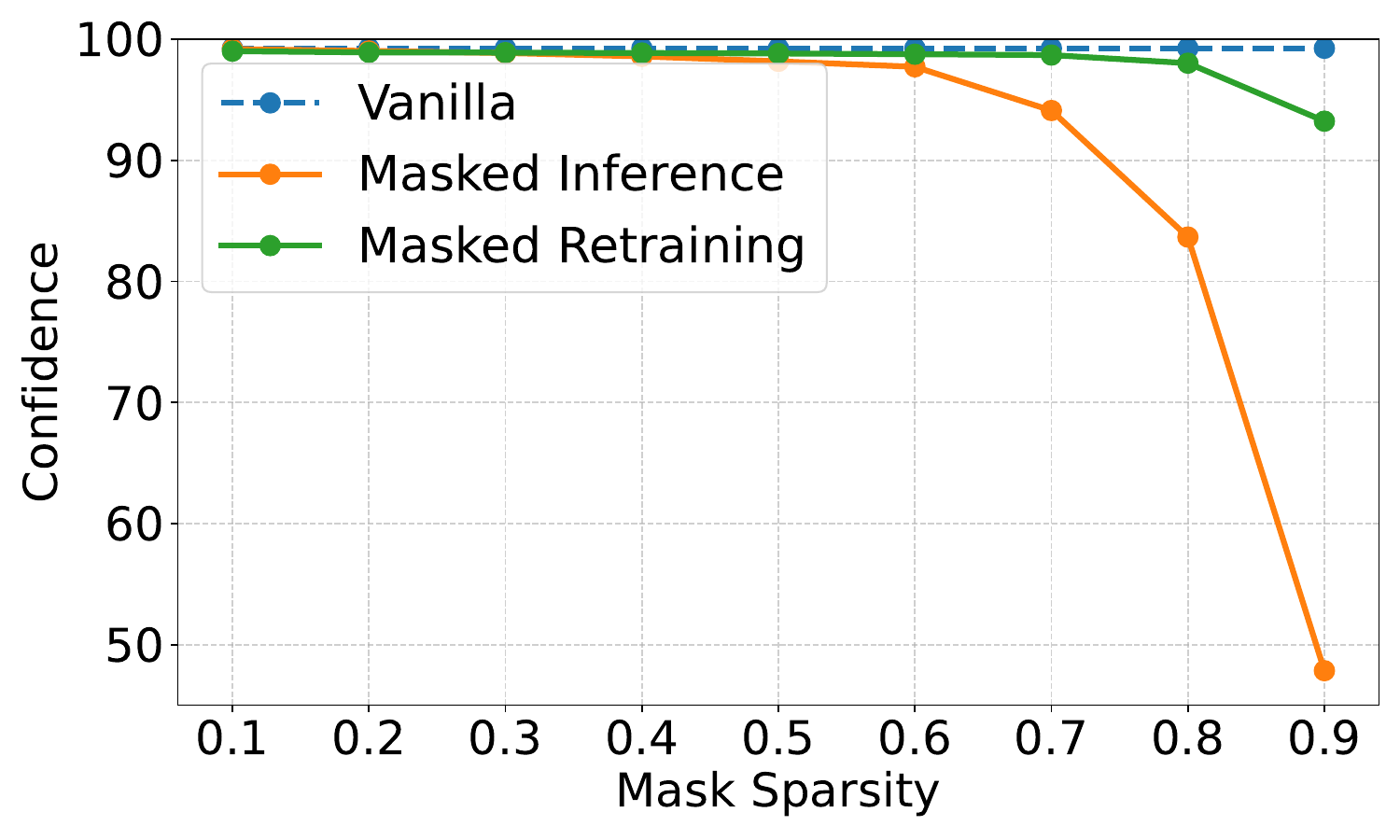}}
  \caption{Post-perturbation statistics. (a) Histogram of prediction confidence across test samples. (b) Classification accuracy under varying masking ratios. (c) Classification confidence under varying masking ratios. \textbf{Masked Inference} applies random parameter masks to the classifier during inference. \textbf{Masked Retraining} applies random parameter masks during classifier retraining with fixed features.}
  \label{fig:1}
\end{figure}

\begin{figure}[htbp]
  \centering
  \subfloat[\label{fig:2a}]{\includegraphics[width=0.33\textwidth]{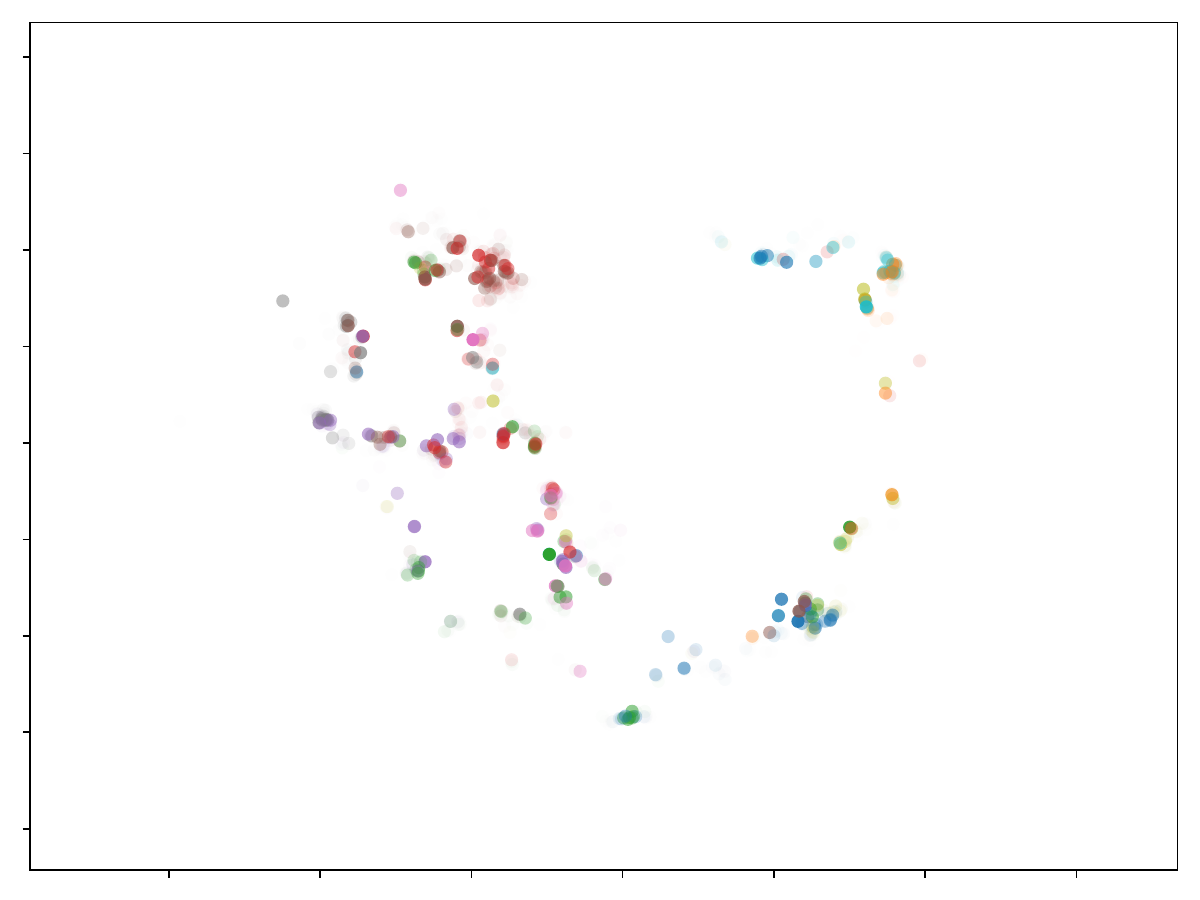}}
  \hfill
  \subfloat[\label{fig:2b}]{\includegraphics[width=0.33\textwidth]{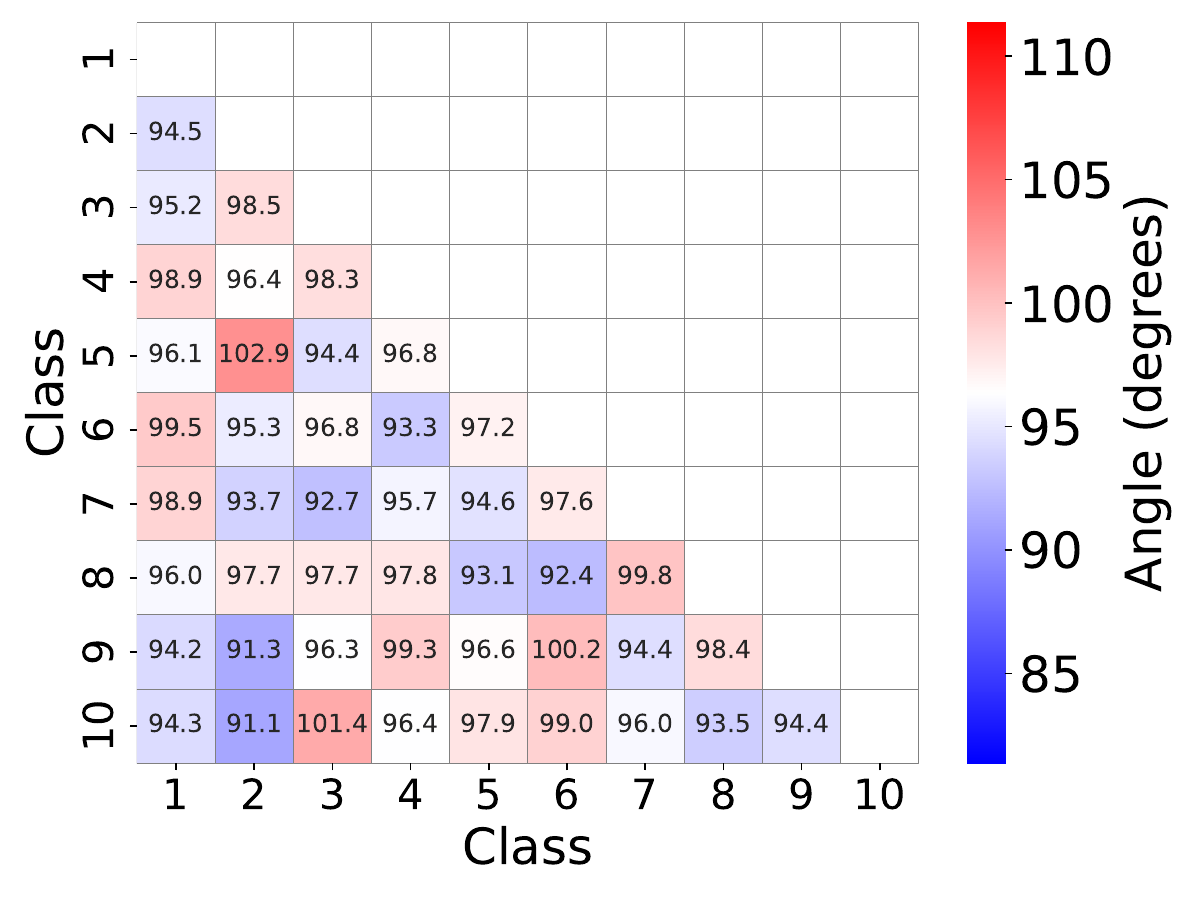}}
  \hfill
  \subfloat[\label{fig:2c}]{\includegraphics[width=0.33\textwidth]{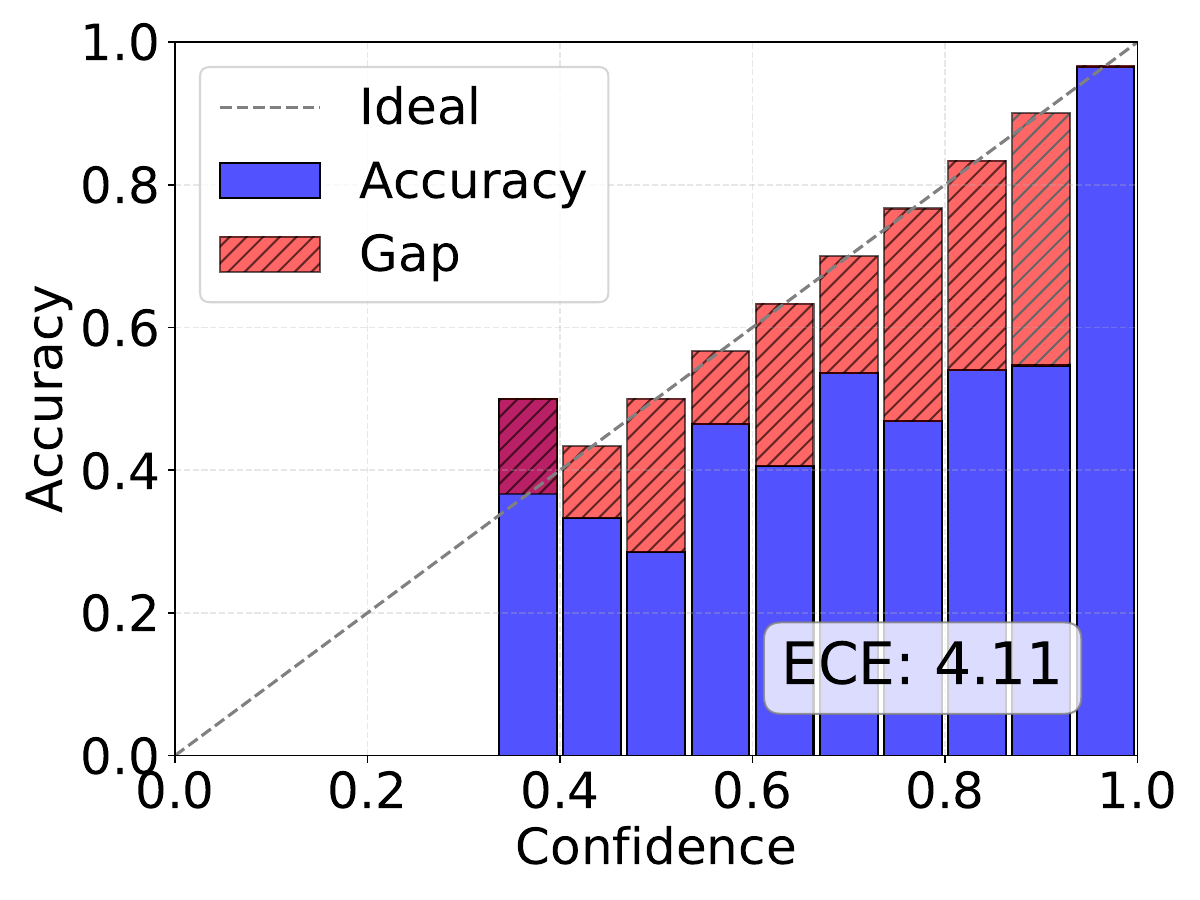}}
  \hfill
  \subfloat[\label{fig:2d}]{\includegraphics[width=0.33\textwidth]{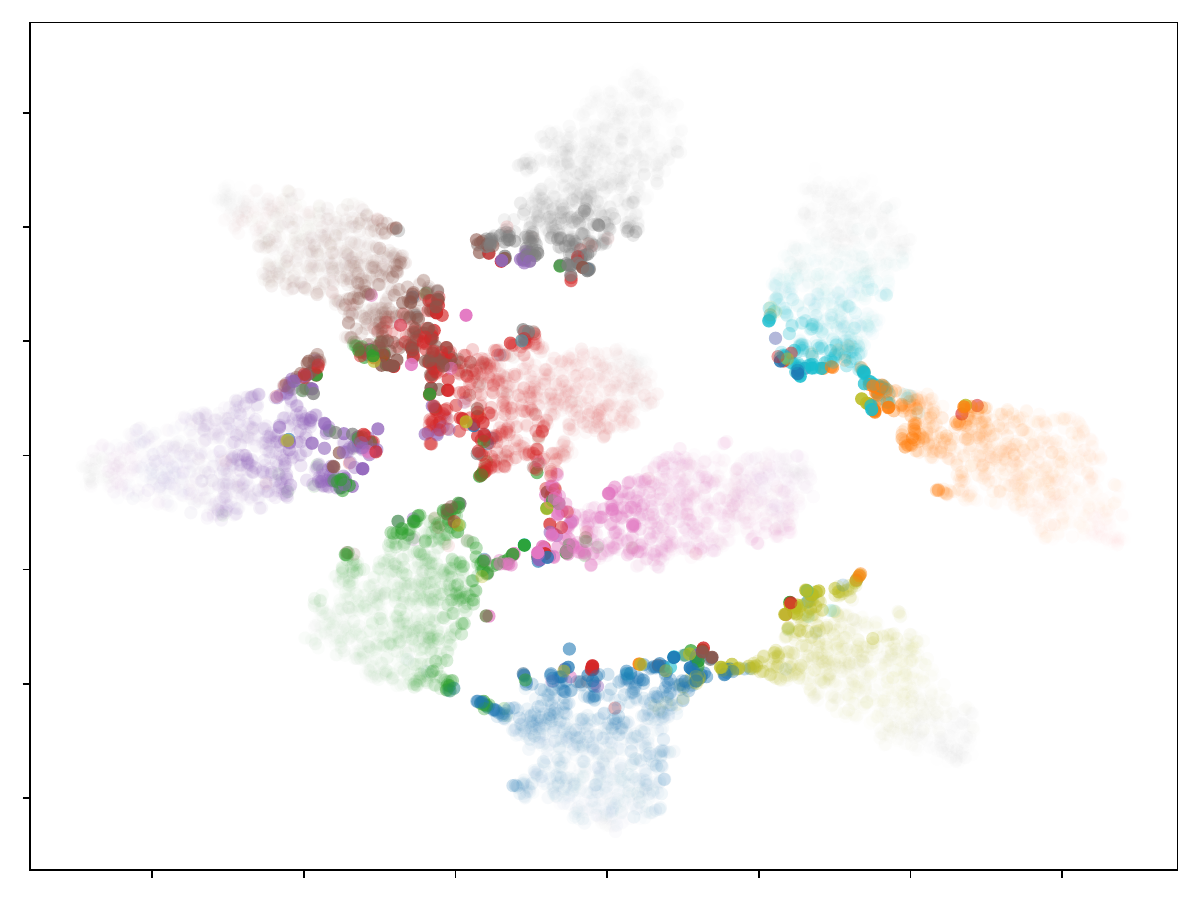}}
  \hfill
  \subfloat[\label{fig:2e}]{\includegraphics[width=0.33\textwidth]{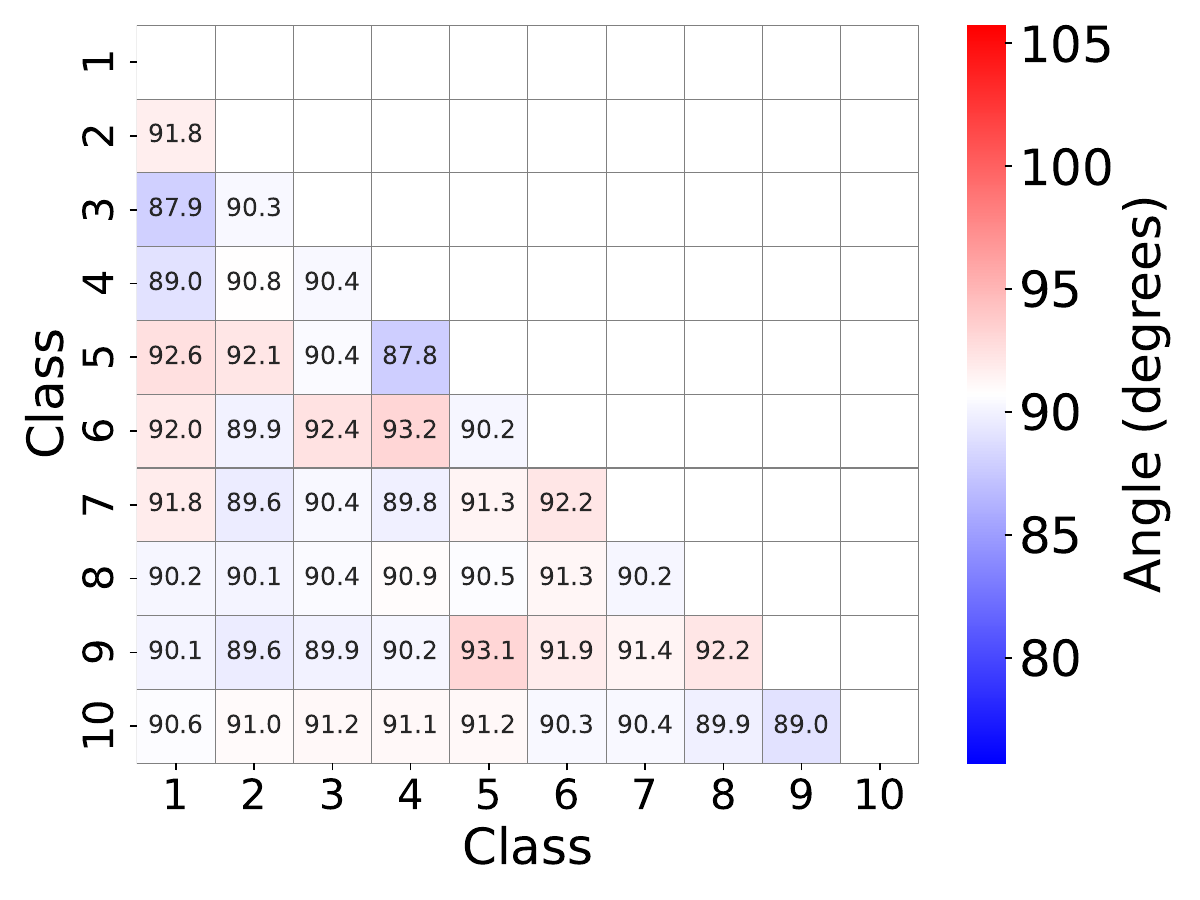}}
  \hfill
  \subfloat[\label{fig:2f}]{\includegraphics[width=0.33\textwidth]{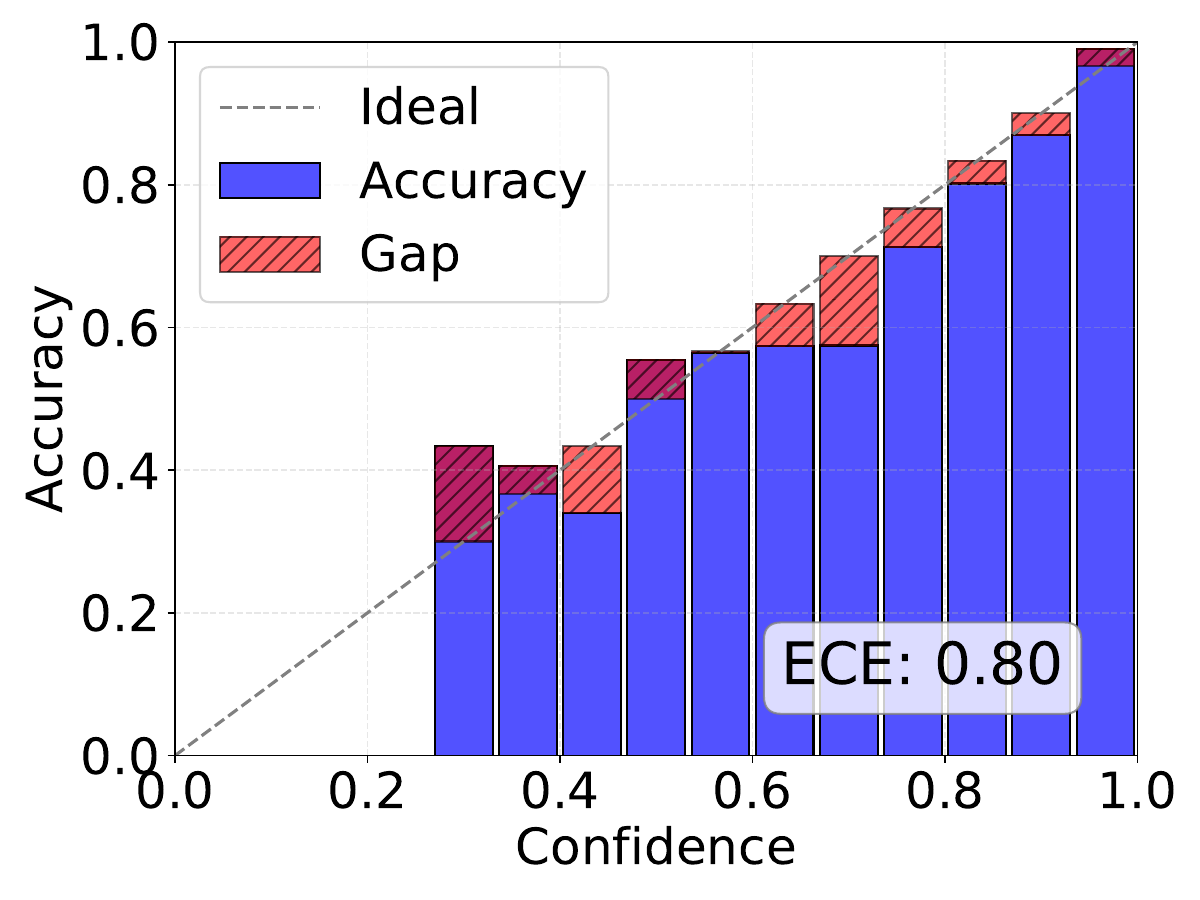}}
  \caption{Visualization results on CIFAR-10 with ResNet-50, illustrating the model before (top) and after (bottom) random mask and retraining. (a) and (d) show the t-SNE visualization of logits, where color denotes class labels and color intensity reflects entropy. (b) and (e) present the angular distance matrix, where lighter values indicate closer proximity to the average angle.
 (c) and (f) display the calibration plots for each setting.}
  \label{fig:2}
\end{figure}

\subsection{Motivation and Analysis}

Previous studies have shown that model miscalibration often stems from overfitting, particularly due to excessive training of the classifier \cite{jordahn2024decoupling,wang2024calibration}. To intuitively demonstrate this miscalibration, we present in Fig.\ref{fig:1a} the confidence distribution of a Vanilla model trained with the cross-entropy loss, where the confidence is primarily concentrated in the high-confidence region. Further analysis of Fig.\ref{fig:1b} and Fig.\ref{fig:1c} reveals that the confidence distribution deviates from the actual accuracy.

We further analyze the causes of the confidence distribution concentration shift. On the one hand, during the late training stage, the feature space gradually stabilizes, and the classifier needs to achieve finer discrimination within a limited representation space, which reduces its sensitivity to boundary samples and uncertain regions. On the other hand, most current approaches to improving model performance focus on optimizing the feature extraction stage to obtain more discriminative and robust feature representations, which generally leads to higher classification accuracy. However, different feature extraction strategies indirectly alter the classifier's response to uncertain samples, thereby influencing its output confidence distribution and ultimately affecting the model's calibration quality.

Taking the cross-entropy loss as example, it is defined as:
\begin{equation}
\mathcal{L}_{\mathrm{CE}} = -\frac{1}{N} \sum_{i=1}^{N} \sum_{k=1}^{K} y_{i,k} \log p_{i,k},
\end{equation}
which essentially maximizes the log-likelihood of the predicted distribution  with respect to the true labels. Although it helps improve classification accuracy, it tends to push model outputs towards extremes (close to 0 or 1), regardless of their alignment with true correctness, leading to overconfidence.

In this context, we raise a key question: can we reshape the confidence expression structure of the classifier to improve model calibration while keeping the feature representation unchanged? To this end, we propose a naive perturbation mechanism: during inference, we fix the feature extractor and randomly mask a subset of classifier parameters at each forward pass. This introduces sparse perturbations to the logits without additional training overhead.
As shown in Fig.~\ref{fig:1a}, this alleviates overconcentration of high-confidence predictions from the vanilla model. However, Figs.~\ref{fig:1b} and \ref{fig:1c} reveal that higher masking rates, while further adjusting confidence, reduce accuracy—indicating a trade-off between calibration and performance.

To address this trade-off, we propose a lightweight  mask-based retraining approach: freezing the feature extractor, we apply random masks to the classifier each epoch and retrain for a few epochs (e.g., 5 epochs). This strategy enables adaptive adjustment of classifier weights, improving calibration while preserving accuracy.
Fig.\ref{fig:2} (a,d) show t-SNE plots of logits colored by ground-truth labels, with color intensity indicating prediction uncertainty (entropy). 
The vanilla model loses most of its uncertainty expression, whereas after mask and retraining, the classifier exhibits clearer boundary structures and better-calibrated confidence. Fig.\ref{fig:2} (b,e) display angular distance matrices of class prototypes, showing more uniform inter-class angles after retraining, indicating a more structured balanced classifier representation. Fig.\ref{fig:2} (c,f) present calibration curves, where the retrained model demonstrates better alignment between confidence and accuracy, indicating more reliable probabilistic estimates.

Overall, the lightweight mask-based retraining serves as a preliminary framework for analyzing the calibration imbalance that arises during training. Building upon this mechanism, we propose a novel method, whose design is presented in Sec.\ref{sec:main_method} and \ref{sec:adaptive}. 
% The effectiveness of each module is validated through ablation studies in Sec.\ref{exp:ablation}.

\begin{figure}[t]
    \centering
    \includegraphics[width=1.0\linewidth]{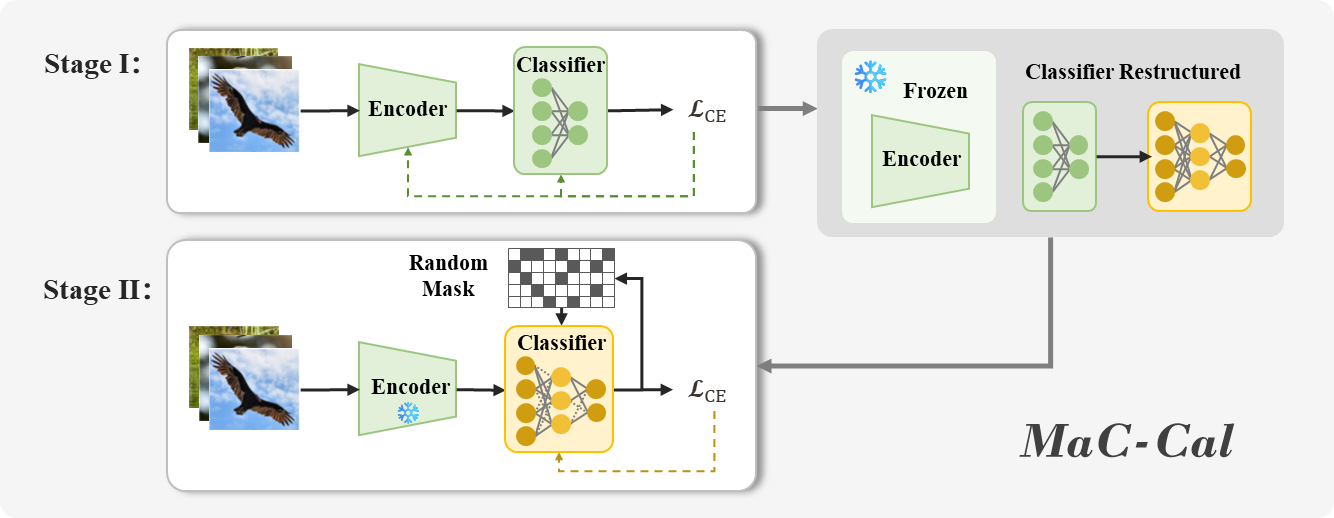}
    \caption{
    Illustration of the MaC-Cal framework. \textbf{Stage I}: joint training of the feature extractor and linear classifier. \textbf{Dark gray (transition phase)}: the feature extractor is frozen and the classifier is restructured. \textbf{Stage II}: masked retraining of the new classifier with random binary masks applied to its weights.
    }
    \label{fig:3}
\end{figure}

\subsection{Mask-based Classifier Calibration}
\label{sec:main_method}
We propose a calibration method for deep neural networks, called \textbf{Mask-based Classifier Calibration (MaC-Cal)}. It introduces stochastic sparsity into the classifier to improve confidence estimation and enhance model calibration.
As shown in Fig.\ref{fig:3}, MaC-Cal adopts a two-stage training strategy. Without altering the task objective or training data, it refines confidence representation via a structurally guided process, showing strong generality and scalability.

\paragraph{\textbf{Classifier Shift Reduction}}
% \noindent\textbf{Classifier Shift Reduction.}
In the first stage, we jointly optimize the feature extractor \( f \) and the classifier \(\mathbf{W}\) by minimizing a standard supervised loss function, such as cross-entropy or focal loss. Data augmentation techniques like mixup can also be applied to enhance generalization. Unless otherwise specified, we adopt cross-entropy loss, original training dataset and train the model until convergence or early stopping is triggered. Formally, the optimization objective is expressed as:
\begin{equation}
\min_{\bm{\theta}, \mathbf{W}} \mathcal{L}_{\mathrm{CE}} \bigl(\{f(x_i; \bm{\theta})\mathbf{W},y_i\}_{i=1}^N\bigr).
\end{equation}

The key of our method is in the second stage, where we freeze the feature extractor parameters \(\bm{\theta}\) learned in the first stage and focus on updating the classifier to further improve the quality of confidence estimation. To enhance the reliability of confidence scores, we introduce a random masking mechanism that applies stochastic binary masks to the classifier weights, inducing variability in classifier behavior. To mitigate potential performance degradation caused by the reduced number of effective parameters—especially in tasks with complex label structures and diverse feature distributions—we replace the original linear classifier with a lightweight two-layer bottleneck structure \(\mathbf{W}_{\text{new}}\), which is randomly initialized. This structure enhances the model’s representational capacity and provides stronger stability during training under sparse perturbations.

Specifically, at each epoch, a random binary mask matrix \(\mathbf{M} \in \{0,1\}^{d \times K}\) is sampled from a Bernoulli distribution:
\begin{equation}
\mathbf{M} \sim \text{Bernoulli}(q).
\label{eq:Mask}
\end{equation}
% The mask affects only the weights and not the biases. 
Each classifier connection is dropped with probability \(1-q\), resulting in sparse connectivity during training. For sample \(i\), the classifier output is:
\begin{equation}
\bm{l}_i = \bm{z}_i (\mathbf{M} \odot \mathbf{W}_{\text{new}}) \in \mathbb{R}^K,
\label{eq:logit}
\end{equation}
where \(\odot\) denotes element-wise multiplication and $\bm{z}_i = f(x_i; \bm{\theta})$. The logits \(\bm{l}_i\) are used to compute the standard cross-entropy loss \(\mathcal{L}_{\text{CE}}\). In the second stage, we only need to optimize classifier $\mathbf{W}_{\text{new}}$ with following optimization objective:
\begin{equation}
\min_{\mathbf{W}_{\text{new}}} \mathcal{L}_{\text{CE}}\bigl(\{\bm{l}_i, y_i\}_{i=1}^N\bigr).
\label{eq:back}
\end{equation}

During backpropagation, the same binary mask \(\mathbf{M}\) is applied to the gradients, zeroing out those corresponding to masked weights to prevent their influence on parameter updates, thereby maintaining training stability. Formally, this update of classifier can be written as:
\begin{equation}
\mathbf{W}'_{\text{new}} =\mathbf{W}_{\text{new}} -\eta*\nabla_{\mathbf{W}_{\text{new}}}, \quad \nabla_{\mathbf{W}_{\text{new}}} = \mathbf{M} \odot \frac{\partial \mathcal{L_{\text{CE}}}}{\partial \mathbf{W}_{\text{new}}},
\label{eq:update}
\end{equation}
where $\eta$ is the learning rate. Designed in this way, the mask matrix influences not only the logit output but also the gradient computation about the classifier.

% \end{minipage}
% \end{center}

\paragraph{\textbf{Adaptive Sparsity}}
\label{sec:adaptive}
During the second stage of MaC-Cal, the sparsity parameter \(q\) governs the generation of the random mask \(\mathbf{M}\), which affects calibration performance. While a fixed \(q\) can work reasonably well, its optimal value varies across models and datasets, making manual tuning inefficient. To overcome this, we propose an adaptive mechanism that adjusts sparsity based on calibration feedback during training.

Specifically, at each training epoch \(t\), we record the model's accuracy \(\text{Acc}_t\) and average confidence \(\text{Conf}_t\) on the training set, and update the mask retention probability \(q_t\) based on their difference. The initial sparsity is set as \(q_0 = 0.5\), and subsequently updated by:
\begin{equation}
q_t = q_{t-1} + \mathrm{clip}\left( \text{Conf}_t - \gamma \cdot \text{Acc}_t,\; -\eta_t,\; \eta_t \right).
\label{eq:adaptive}
\end{equation}

In practical applications where validation or test data are often scarce or unavailable, the method operates directly on training-set metrics to enable feedback-driven adjustment.
However, since models tend to overfit the training data during optimization, the observed training accuracy often overestimates true generalization. To mitigate this, \(\gamma \in (0,1)\) serves as a down-weighting factor that scales back the influence of over-optimistic accuracy and approximates performance on unseen data.
To ensure stable and gradual updates, we apply a clipped adjustment bounded by a dynamic threshold \(\eta_t\), which decays exponentially as training progresses:
\begin{equation}
\eta_t = \eta_{\text{init}} \cdot \exp\left( \log\left( \frac{\eta_{\text{final}}}{\eta_{\text{init}}} \right) \cdot \frac{t}{T} \right),
\label{eq:eta_progress}
\end{equation}
where \(\eta_{\text{init}} = 0.1\), \(\eta_{\text{final}} = 0.001\), and \(T\) is the total number of epochs. This schedule allows large, flexible corrections at early stages and more conservative refinement later, preventing oscillations or overshooting. By leveraging accuracy-confidence feedback on the training set, while compensating for its inherent optimism, \textbf{MaC-Cal} enables robust and adaptive control over mask sparsity throughout training.

\begin{algorithm}[t]
    \centering
    \renewcommand{\algorithmicrequire}{\textbf{Input:}}
    \renewcommand{\algorithmicensure}{\textbf{Output:}}
    \caption{Second-Stage Training with MaC-Cal}
    \label{alg}
    \begin{algorithmic}[1]
        \REQUIRE Training dataset \( D \), frozen feature extractor \( f(x; \bm{\theta}) \), randomly initialized classifier \(\mathbf{W_{new}}\), hyperparameters.
        \ENSURE Updated classifier \(\mathbf{W_{new}}\)
        \STATE Initialize mask sparsity \(q_1 \leftarrow 0.5\);
        \FOR{$t = 1, 2, \ldots, T$}
            \FOR{each mini-batch $\{x, y\} \in D$}
                \STATE Sample mask matrix \(\mathbf{M}\) from Eq.~(\ref{eq:Mask}) with corresponding $q_t$;
                \STATE Extract feature vector \(\bm{z} \leftarrow f(x; \bm{\theta})\);
                \STATE Compute logit \(\bm{l}\) via Eq.~(\ref{eq:logit}) and CE loss \(\mathcal{L_{\text{CE}}}\) via Eq.(\ref{eq:back});
                \STATE Update \(\mathbf{W_{new}}\) as $\mathbf{W}_{\text{new}} -\eta*\nabla_{\mathbf{W}_{\text{new}}}$ via masked gradient backpropagation Eq.(\ref{eq:update});
            \ENDFOR
            \STATE Compute \(\text{Acc}_t\) and \(\text{Conf}_t\);
            \STATE Update \( q_t \) via Eq.~(\ref{eq:adaptive}) and clamp \( q_t \) to valid range \([0, 1]\) if necessary;
        \ENDFOR
    \end{algorithmic}
\end{algorithm}
\subsection{Discussions}
The proposed MaC-Cal framework addresses miscalibration in a principled manner by decoupling feature learning from confidence estimation. 
It employs a two-stage training scheme: in the first stage, the feature extractor is optimized to produce discriminative representations; in the second stage, the classifier is adapted through random masking and adaptive retraining, encouraging the model to better align prediction confidence with accuracy. The detailed procedure of the second stage is summarized in Algorithm \ref{alg}.

Our method is straightforward to implement and can be readily integrated into existing training pipelines with minimal computational overhead. In contrast to regularization techniques such as dropout, which randomly drops neurons during training, MaC-Cal introduces structured sparsity at the classifier weight level and further incorporates this sparsity into gradient computation. This design ensures deterministic behavior during inference. Additionally, we propose an adaptive sparsity mechanism that replaces fixed dropout rates, mitigating instability caused by manual hyperparameter tuning. Consequently, our approach produces more stable confidence estimates and enhances calibration reliability.

In addition to the advantages discussed above, MaC-Cal not only addresses overconfidence but also effectively mitigates underconfidence, a critical yet often overlooked source of calibration error. This dual capability distinguishes MaC-Cal as a robust calibration method capable of producing reliable confidence estimates across varying model behaviors. A representative example of underconfidence occurs when training with Mixup—a data augmentation technique that linearly interpolates between pairs of inputs. Due to the softened decision boundaries introduced by these interpolated samples, models trained with Mixup often yield underconfident predictions. Formally, Mixup generates virtual samples as:
\begin{equation}
x = \lambda x_i + (1 - \lambda) x_j, \quad \tilde{y} = \lambda y_i + (1 - \lambda) y_j,
\end{equation}
where $\lambda \sim \mathrm{Beta}(\alpha, \alpha)$. Although Mixup improves generalization, it can adversely affect confidence estimation during inference. Specifically, calibration performance is highly sensitive to the choice of the mixing coefficient $\alpha$. Smaller values of $\alpha$ tend to improve calibration, whereas larger values generally enhance generalization but often lead to underconfident predictions and increased ECE. 
To mitigate this, our method applies Mixup in the first stage to learn a robust feature extractor, then trains the classifier on original (non-augmented) data during the second stage to obtain clearer decision boundaries. This two-stage approach improves calibration while maintaining accuracy and robustness against variations in $\alpha$.

\section{Experiments}
\label{Experiment}

\subsection{Implementation Details}

\paragraph{\textbf{Datasets}}
We evaluate our method on several architectures, including ResNet-50, ResNet-110 \cite{He_2016_CVPR}, WRN-26-10 \cite{zagoruyko2016wide}, and DenseNet-121 \cite{huang2017densely}, across CIFAR-10/100 \cite{krizhevsky2009learning}, and Tiny-ImageNet \cite{le2015tiny} for calibration performance. 
To comprehensively assess model robustness, we conduct additional evaluations on SVHN~\cite{netzer2011reading} as an OOD dataset, and on CIFAR-10-C and CIFAR-100-C~\cite{hendrycks2019benchmarking}, which include various types of synthetic noise and input perturbations, effectively simulating diverse real-world distributional changes in a more comprehensive manner.

\paragraph{\textbf{Evaluation Protocols}}
To evaluate calibration, we report ECE, AECE~\cite{nguyen2015posterior}, and MCE. We also report top-1 accuracy to assess classification performance. For OOD detection, we adopt AUROC~\cite{bradley1997use} and FPR95 to assess the model's ability to distinguish in-distribution and out-of-distribution samples. These metrics jointly reflect the reliability and robustness of model predictions.

\paragraph{\textbf{Methods for Comparison}}

We compare our method against a variety of established calibration approaches. These include the original baseline model trained with cross-entropy loss, implicit calibration methods such as ACLS \cite{park2023acls}, MbLS\cite{liu2022devil}, Label Smoothing \cite{szegedy2016rethinking}, Mixup \cite{thulasidasan2019mixup}, MIT \cite{wang2023pitfall}, Focal Loss \cite{mukhoti2020calibrating}, and FLSD \cite{mukhoti2020calibrating}, as well as explicit calibration methods including CPC \cite{Cao2019}, PLP \cite{wang2024calibration}, MMCE \cite{kumar2018trainable}, TST/VTST \cite{jordahn2024decoupling}, and BalCAL \cite{ni2025balancing}. 
For methods with publicly available code and hyperparameters, we strictly follow the original configurations during reproduction. 

\paragraph{\textbf{Training Details}}
Our training protocol follows the setup in \cite{mukhoti2020calibrating}, using their publicly released codebase as the foundation.  
On CIFAR-10/100, models train for 350 epochs with 5,000 images for validation; learning rate decays from 0.1 to 0.01 after 150 epochs, then to 0.001 after 250 epochs. Tiny-ImageNet training lasts 100 epochs with a similar learning rate schedule (0.1 for 40 epochs, 0.01 for 20, 0.001 for 40). For two-stage methods (TST/VTST~\cite{jordahn2024decoupling} and ours), the second stage adds 40 epochs at 0.1 LR. We use SGD with momentum 0.9, weight decay $5 \times 10^{-4}$, batch size 128, on one NVIDIA A40 GPU with fixed seed 1 for reproducibility.

\begin{table}[t]
  \centering
  \caption{ECE of different methods evaluated across three datasets and four models.}
  \resizebox{1.0\textwidth}{!}{%
  \setlength{\tabcolsep}{2pt}
    \begin{tabular}{c|cccc|cccc|c}
    \toprule
    \multirow{2}[4]{*}{\textbf{Method}} & \multicolumn{4}{c|}{ CIFAR-10} & \multicolumn{4}{c|}{ CIFAR-100} & Tiny-ImageNet \\
\cmidrule{2-10}          & ResNet-50 & ResNet-110 & WRN-26-10 & DenseNet-121 & ResNet-50 & ResNet-110 & WRN-26-10 & DenseNet-121 & ResNet-50 \\
    \midrule
    \textbf{Vanilla} & 4.11  & 4.06  & 4.02  & 3.68  & 14.36  & 17.85  & 11.74  & 16.04  & 5.53  \\
    \textbf{ACLS\cite{park2023acls}} & 2.40  & 2.56  & 1.58  & 2.55  & 7.94  & 12.28  & 8.93  & 13.43  & 2.37  \\
    \textbf{LS\cite{szegedy2016rethinking}} & 3.96  & 3.45  & 4.01  & 3.21  & 4.28  & 4.40  & 3.46  & 8.83  & 2.51  \\
    \textbf{MbLS\cite{liu2022devil}} & 1.37  & 1.54  & 1.46  & 1.69  & 6.66  & 11.10  & 8.28  & 11.88  & 10.17  \\
    \textbf{mixup\cite{thulasidasan2019mixup}} & 1.74  & 4.27  & 1.35  & 2.25  & 4.82  & 5.40  & 4.45  & 4.80  & 3.26  \\
    \textbf{MIT\cite{wang2023pitfall}} & 2.61  & 3.14  & 2.40  & 2.89  & 7.09  & 11.71  & 5.35  & 8.38  & 1.86  \\
    \textbf{FL\cite{mukhoti2020calibrating}} & 2.34  & 2.09  & 1.70  & 2.06  & 5.56  & 7.80  & 3.03  & 4.65  & 1.99  \\
    \textbf{FLSD\cite{mukhoti2020calibrating}} & 2.50  & 2.42  & 1.24  & 2.42  & 5.46  & 7.83  & 3.21  & 4.42  & 2.06  \\
    \textbf{DFL\cite{tao2023dual}} & 2.21  & 2.08  & 1.47  & 1.56  & 3.01  & 4.69  & 2.98  & 2.08  & 5.06  \\
    \textbf{CPC\cite{Cao2019}} & 3.92  & 4.00  & 5.42  & 3.70  & 7.42  & 12.73  & 7.81  & 8.97  & 9.17  \\
    \textbf{MMCE\cite{kumar2018trainable}} & 2.71  & 3.25  & 2.68  & 1.90  & 6.00  & 11.37  & 4.56  & 6.79  & 7.55  \\
    \textbf{PLP\cite{wang2024calibration}} & 2.97  & 3.95  & 2.34  & 2.86  & 7.73  & 12.95  & 7.89  & 8.19  & 11.07  \\
    \textbf{TST\cite{jordahn2024decoupling}} & 4.26  & 4.11  & 4.09  & 4.67  & 11.01  & 16.82  & 8.06  & 13.51  & 5.21  \\
    \textbf{VTST\cite{jordahn2024decoupling}} & 4.41  & 4.31  & 3.97  & 4.68  & 4.83  & 14.62  & 4.39  & 8.73  & 7.20  \\
    \textbf{BalCAL\cite{ni2025balancing}} & 1.32  & 1.51  & 1.45  & 1.46  & 3.45  & 3.91  & 3.07  & 2.94  & 1.54  \\
    \midrule
    \textbf{Ours} & \textbf{0.92} & \textbf{1.25} & \textbf{0.96} & \textbf{1.05} & \textbf{2.90} & \textbf{1.44} & \textbf{2.96} & \textbf{1.73} & \textbf{1.30} \\
    \bottomrule
    \end{tabular}%
 }
  \label{tab:main}%
\end{table}%

\begin{figure}[t]
    \centering
    \includegraphics[width=1.0\linewidth]{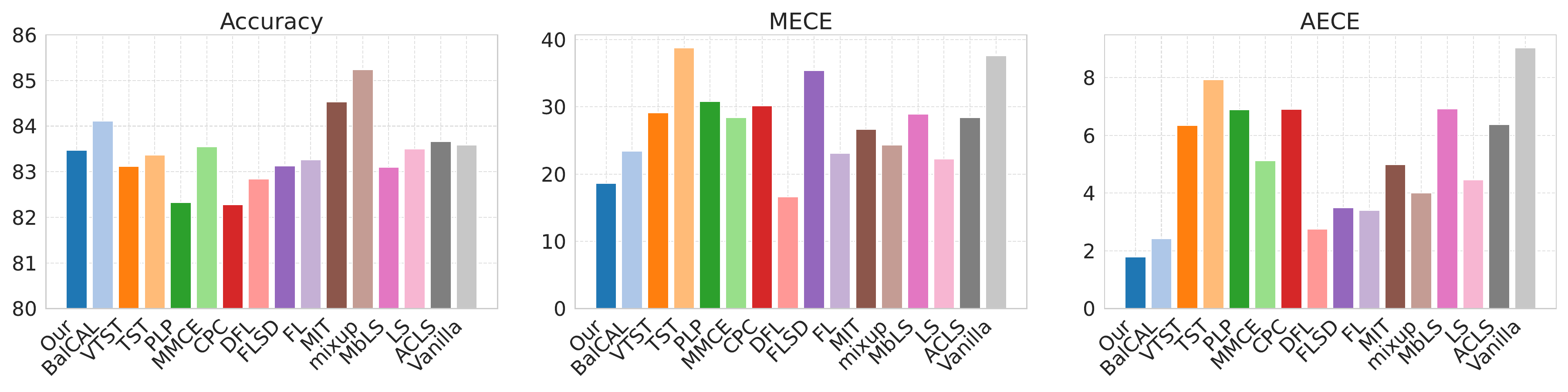}
    \caption{Average performance metrics across all models and datasets.}
    \label{fig:main}
\end{figure}

\subsection{Main Results}
% To comprehensively evaluate the calibration performance of different methods, we systematically compare 15 representative calibration methods on four widely used network architectures, focusing on the ECE. As shown in Tab.\ref{tab:main}, the proposed method consistently demonstrates superior performance across all experimental settings. For instance, on the ResNet-110 model trained on CIFAR-100, the ECE is significantly reduced from 17.85 in the baseline model to 1.44, achieving a relative reduction exceeding 90\%.

% Furthermore, we compute the average classification accuracy, MCE, and AECE across all models and datasets. As summarized in Fig.\ref{fig:main}, the proposed method maintains competitive accuracy while substantially reducing both MCE and AECE, demonstrating more reliable confidence estimation without sacrificing predictive performance.

To comprehensively evaluate the performance of various calibration methods, we systematically compare 15 representative approaches across four widely used network architectures, focusing on the ECE. As shown in Tab.~\ref{tab:main}, the proposed method consistently achieves the best performance across all models and datasets, demonstrating both superior stability and significant improvements over existing methods. For example, on CIFAR-100 with ResNet-110, the baseline model yields an ECE of 17.85, while our method reduces it to 1.44, achieving over 90\% relative reduction. 
% Similarly, on CIFAR-10 and Tiny-ImageNet, our method attains the lowest ECE across all models, indicating strong adaptability to both simple and complex tasks.

Furthermore, we compute the average classification accuracy, MCE, and AECE across all datasets and models, as summarized in Fig.~\ref{fig:main}. Compared to other methods, our approach effectively reduces both MCE and AECE while maintaining competitive accuracy. This indicates that our method not only improves the consistency of overall confidence estimation but also reduces the worst-case confidence error, resulting in more reliable calibration quality.

\begin{table}[t]
  \centering
  \caption{Performance comparison of OOD detection.}
  \resizebox{0.55\textwidth}{!}{%
    \begin{tabular}{c|cc|cc}
      \toprule
            & \multicolumn{2}{c|}{CIFAR-10} & \multicolumn{2}{c}{CIFAR-100} \\
            & AUROC↑ & FPR95↓ & AUROC↑ & FPR95↓ \\
      \midrule
      Vanilla & 88.07  & 56.79  & 78.66  & 82.91  \\
      PLP   & \textbf{91.53}  & \textbf{49.66}  & 77.29  & 79.13  \\
      TST   & 87.24  & 57.75  & 78.41  & 81.52  \\
      VTST  & 83.54  & 59.49  & 78.50  & 80.18  \\
      BalCAL & 88.85 & 54.75  & \underline{78.95}  & \underline{78.99}  \\
      \midrule
      Ours   & \underline{88.91}  & \underline{54.11}  & \textbf{79.05}  & \textbf{77.06}  \\
      \bottomrule
    \end{tabular}
 }
    \label{tab:ood}
\end{table}

\begin{figure}[t]
    \centering
    \includegraphics[width=1.0\linewidth]{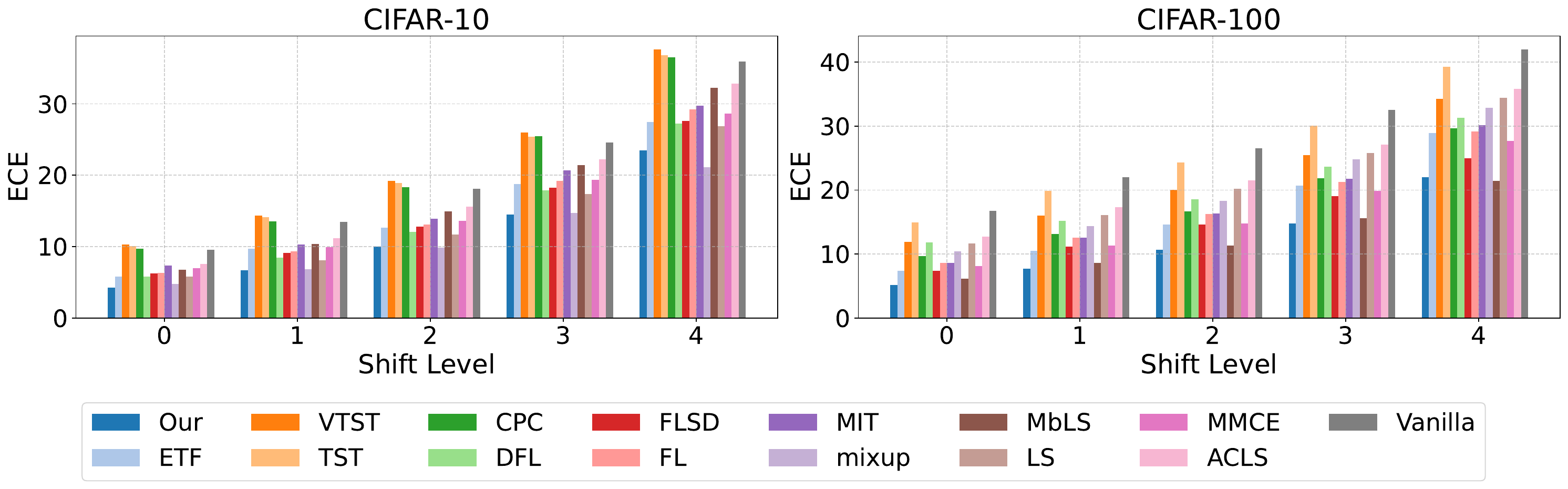}
    \caption{Calibration performance under distribution shifts on different datasets.}
    \label{fig:shift}
\end{figure}

\subsection{Robustness Evaluation}

To assess robustness under perturbations and distribution shifts, we evaluate models trained on CIFAR-10 and CIFAR-100 using their corrupted versions (CIFAR-10-C and CIFAR-100-C), including blur, noise, compression artifacts, and lighting changes. Fig.~\ref{fig:shift} shows that while most methods suffer under severe corruptions, MaC-Cal consistently achieves relatively low ECE values, indicating better robustness and more reliable confidence estimates in challenging conditions.

We further assess generalization on OOD data: models trained on CIFAR-10 are tested on CIFAR-100 and SVHN, and vice versa. Tab.~\ref{tab:ood} reports the average performance. MaC-Cal consistently outperforms the Vanilla and ranks among the top-performing methods, showing strong generalization.

\subsection{Mixup and Underconfidence}
\begin{figure}[t]
    \centering
    \includegraphics[width=0.8\linewidth]{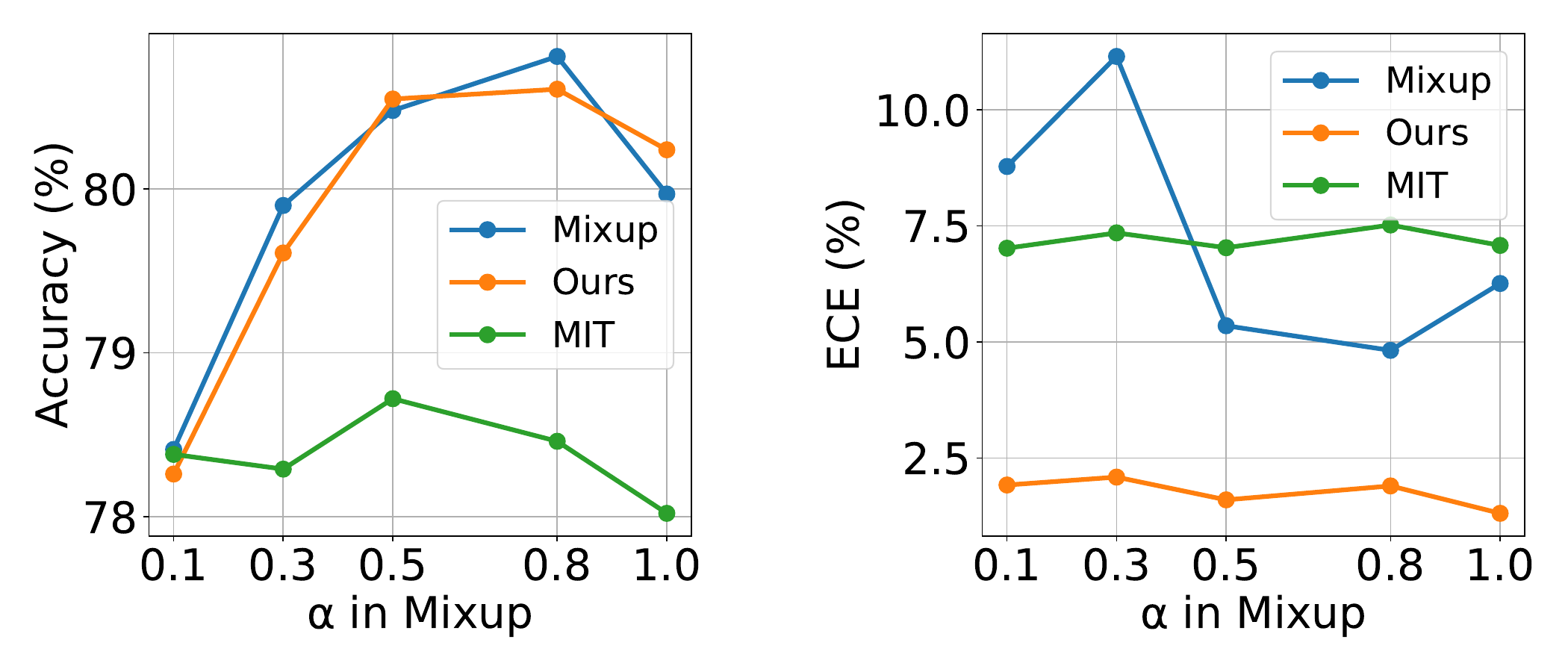}
    \caption{Results under varying mixing parameters $\alpha$ on ResNet-50 and CIFAR-10.}
    \label{fig:mixup}
\end{figure}

In our experiments, Mixup \cite{thulasidasan2019mixup} is adopted as one of the baselines for model calibration. While it improves generalization and accuracy, its calibration performance is highly sensitive to the mixing parameter $\alpha$. Large $\alpha$ values tend to produce underconfident predictions and may increase ECE, despite mitigating overfitting. Thus, choosing $\alpha$ involves balancing accuracy and calibration.
To address this, Mixup Inference in Training (MIT) \cite{wang2023pitfall} introduces a decoupling strategy that recovers raw outputs at the end of the forward pass to improve calibration. While effective, MIT significantly increases computational cost, especially for large models. We compare Mixup, MIT, and our method under varying $\alpha$ (Fig.~\ref{fig:mixup}). 
Both MIT and our method reduce ECE, but ours achieves better calibration while preserving Mixup's accuracy gains. Moreover, our method is more efficient than MIT, as detailed in Sec.\ref{sec:cost}.

\begin{table}[t]
  \centering
  \caption{Performance of baseline calibration methods with and without our method.}
  \resizebox{1.0\textwidth}{!}{%
    \begin{tabular}{c|cccc|cccc|cccc}
% Tab generated by Excel2LaTeX from sheet 'Sheet8'
    \toprule
    \multirow{3}[3]{*}{Method} & \multicolumn{4}{c|}{CIFAR-10} & \multicolumn{4}{c|}{CIFAR-100} & \multicolumn{4}{c}{Tiny-ImageNet} \\
          & \multicolumn{2}{c}{baseline} & \multicolumn{2}{c|}{+Ours} & \multicolumn{2}{c}{baseline} & \multicolumn{2}{c|}{+Ours} & \multicolumn{2}{c}{baseline} & \multicolumn{2}{c}{+Ours} \\
\cmidrule{2-13}          
      & \textcolor{darkgray}{Acc↑}   & ECE↓  
      & \textcolor{darkgray}{Acc↑}   & ECE↓    
      & \textcolor{darkgray}{Acc↑}   & ECE↓  
      & \textcolor{darkgray}{Acc↑}   & ECE↓    
      & \textcolor{darkgray}{Acc↑}   & ECE↓  
      & \textcolor{darkgray}{Acc↑}   & ECE↓ \\
    \midrule

    ACLS  & \textcolor{darkgray}{\textbf{95.23}} & 2.27 & \textcolor{darkgray}{95.22} & \textbf{1.44} & \textcolor{darkgray}{77.74} & 10.65 & \textcolor{darkgray}{\textbf{77.77}} & \textbf{3.49} & \textcolor{darkgray}{\textbf{61.04}} & 2.36 & \textcolor{darkgray}{60.77} & \textbf{1.15} \\
    LS    & \textcolor{darkgray}{\textbf{95.09}} & 3.66 & \textcolor{darkgray}{95.07} & \textbf{1.41} & \textcolor{darkgray}{\textbf{77.54}} & 5.24 & \textcolor{darkgray}{77.52} & \textbf{3.47} & \textcolor{darkgray}{60.98} & 2.50 & \textcolor{darkgray}{\textbf{61.27}} & \textbf{1.17} \\
    MbLS  & \textcolor{darkgray}{95.18} & 1.52 & \textcolor{darkgray}{\textbf{95.20}} & \textbf{1.35} & \textcolor{darkgray}{77.65} & 9.48 & \textcolor{darkgray}{\textbf{77.68}} & \textbf{3.74} & \textcolor{darkgray}{56.58} & 10.15 & \textcolor{darkgray}{\textbf{56.64}} & \textbf{0.85} \\
    mixup & \textcolor{darkgray}{\textbf{96.12}} & 2.40 & \textcolor{darkgray}{96.03} & \textbf{1.10} & \textcolor{darkgray}{\textbf{80.82}} & 4.86 & \textcolor{darkgray}{80.75} & \textbf{3.29} & \textcolor{darkgray}{60.06} & 7.52 & \textcolor{darkgray}{\textbf{61.50}} & \textbf{0.62} \\
    MIT   & \textcolor{darkgray}{96.15} & 2.76 & \textcolor{darkgray}{\textbf{96.26}} & \textbf{0.92} & \textcolor{darkgray}{\textbf{78.67}} & 8.05 & \textcolor{darkgray}{78.43} & \textbf{2.75} & \textcolor{darkgray}{61.72} & 1.87 & \textcolor{darkgray}{\textbf{61.91}} & \textbf{0.76} \\
    FL    & \textcolor{darkgray}{\textbf{95.16}} & 2.05 & \textcolor{darkgray}{94.98} & \textbf{0.71} & \textcolor{darkgray}{77.44} & 5.26 & \textcolor{darkgray}{\textbf{77.67}} & \textbf{1.70} & \textcolor{darkgray}{58.99} & 1.98 & \textcolor{darkgray}{\textbf{59.27}} & \textbf{0.92} \\
    FLSD  & \textcolor{darkgray}{95.00} & 2.15 & \textcolor{darkgray}{\textbf{95.14}} & \textbf{0.81} & \textcolor{darkgray}{\textbf{77.36}} & 5.23 & \textcolor{darkgray}{76.45} & \textbf{1.36} & \textcolor{darkgray}{58.69} & 2.06 & \textcolor{darkgray}{\textbf{59.24}} & \textbf{0.85} \\
    DFL   & \textcolor{darkgray}{\textbf{95.09}} & 1.83 & \textcolor{darkgray}{94.98} & \textbf{0.64} & \textcolor{darkgray}{\textbf{77.63}} & 3.15 & \textcolor{darkgray}{77.17} & \textbf{1.35} & \textcolor{darkgray}{54.65} & 5.06 & \textcolor{darkgray}{2.76} & \textbf{0.80} \\
    CPC   & \textcolor{darkgray}{93.68} & 4.26 & \textcolor{darkgray}{\textbf{94.33}} & \textbf{1.35} & \textcolor{darkgray}{77.42} & 9.23 & \textcolor{darkgray}{\textbf{77.46}} & \textbf{5.35} & \textcolor{darkgray}{56.12} & 9.17 & \textcolor{darkgray}{\textbf{57.52}} & \textbf{5.74} \\
    MMCE  & \textcolor{darkgray}{95.11} & 2.64 & \textcolor{darkgray}{\textbf{95.18}} & \textbf{0.98} & \textcolor{darkgray}{\textbf{77.48}} & 7.18 & \textcolor{darkgray}{77.21} & \textbf{2.06} & \textcolor{darkgray}{\textbf{61.57}} & 7.55 & \textcolor{darkgray}{60.43} & \textbf{1.34} \\

    \bottomrule
    \end{tabular}%
   }
  \label{tab:combination}%

\end{table}

\subsection{Combinations with Other Methods}
As a plug-and-play method, ours can be seamlessly integrated with a wide range of existing calibration techniques. Tab.~\ref{tab:combination} presents the average calibration results across multiple neural network architectures, comparing the original baseline methods with their enhanced versions combined with our approach. Since TST/VTST \cite{jordahn2024decoupling}, PLP \cite{jordahn2024decoupling}, and BalCAL \cite{ni2025balancing} focus specifically on classifier-level calibration, similar to the core mechanism of our method, we exclude them from combination experiments to avoid overlapping effects and ensure a fair evaluation.

Across various settings, integrating our method consistently reduces ECE while maintaining accuracy. For example, combined with ACLS, our method reduces ECE drops from 2.27 to 1.44 on CIFAR-10 and from 10.65 to 3.49 on CIFAR-100, showing better alignment between confidence and accuracy. These results confirm the broad applicability and effectiveness of our method in improving model calibration.

\begin{table}[t]
  \centering
  \caption{Computational cost in seconds for the entire training process.}
  \resizebox{0.85\textwidth}{!}{%
    \begin{tabular}{c|c|cccccc}
    \toprule
     Dataset &  Model & \textcolor{darkgray}{Vanilla} & MIT   & TST   & VTST  & BalCAL & Ours \\
    \midrule
    \multirow{4}[1]{*}{CIFAR-10} & ResNet-50 & \textcolor{darkgray}{13691} & 25172  & 14959  & 14361  & 15334  & \textbf{14358}  \\
          & ResNet-110 & \textcolor{darkgray}{22940} & 42819  & 24312  & 24001  & 24613  & \textbf{24003}  \\
          & WRN-26-10 & \textcolor{darkgray}{20754} & 39012  & 22633  & 21697  & 22302  & \textbf{21678}  \\
          & DenseNet-121 & \textcolor{darkgray}{16318} & 30354  & 17843  & 17104  & 17795  & \textbf{17102}  \\
          \midrule
    \multirow{4}[0]{*}{ CIFAR-100} & ResNet-50 & \textcolor{darkgray}{13726} & 24358  & 15190  & 14418  & 15746  & \textbf{14417}  \\
          & ResNet-110 & \textcolor{darkgray}{22962} & 42711  & 24047  & 24027  & 25265  & \textbf{24024}  \\
          & WRN-26-10 & \textcolor{darkgray}{20619} & 38402  & 22309  & 21602  & 22955  & \textbf{21554}  \\
          & DenseNet-121 & \textcolor{darkgray}{16702} & 30297  & 18157  & \textbf{17521}  & 18216  & 17572  \\
          \midrule
    Tiny-ImageNet & ResNet-50 & \textcolor{darkgray}{106914} & 204046  & 112040  & 112042  & 119791  & \textbf{112028}  \\
    \bottomrule
    \end{tabular}%
   }
  \label{tab:time}%
\end{table}%

\subsection{Computational Cost}
\label{sec:cost}
% Although the vanilla model has the lowest training time due to the absence of any calibration, it cannot serve as a meaningful baseline for calibration performance. 
To fairly evaluate the computational cost of our method, we compare total training time with other calibration techniques that introduce additional overhead, as summarized in Tab.\ref{tab:time}. Please note that while the vanilla model is the fastest due to the absence of calibration, it does not serve as a meaningful baseline for evaluating calibration performance. Among calibration methods, MaC-Cal consistently shows the lowest or near-lowest training time. On CIFAR-10/100 with ResNet and DenseNet, it outperforms MIT and BalCAL in efficiency, and matches or slightly improves over TST and VTST while delivering better calibration. This shows MaC-Cal offers an efficient trade-off between cost and performance in resource-limited settings.

\begin{table*}[t]
  \centering
  \caption{Effect of MaC-Cal components on ResNet59 performance.}
  \resizebox{1.0\textwidth}{!}{%
    \begin{tabular}{c|cccc|cccc}
    \toprule
     \multirow{2}[4]{*}{Component} & \multicolumn{4}{c|}{CIFAR-10}  & \multicolumn{4}{c}{CIFAR-100} \\
\cmidrule{2-9}          & Acc   & ECE   & AECE  & MCE   & Acc   & ECE   & AECE  & MCE \\
    \midrule
    Vanilla & 94.61  & 3.68  & 3.68  & 29.64  & \textbf{78.03} & 16.04  & 16.03  & 47.42  \\
    +Masked Retraining & 94.66  & 3.45  & 3.42  & 25.74  & 77.80  & 13.11  & 13.08  & 35.69  \\
    +Gradient Restriction & 94.28  & 1.96  & 1.95  & 75.02  & 76.15  & 2.94  & 2.86  & 9.84  \\
    +Classifier Restructure & 95.23  & 1.96  & 1.89  & 26.25  & 77.95  & 2.22  & 2.51  & 10.75  \\
    +Static Adaptive Sparsity & 95.23  & 1.44  & 1.39  & 18.64  & 77.24  & 1.81  & 1.96  & \textbf{7.42} \\
    +Decaying Adaptive Sparsity & \textbf{95.26} & \textbf{1.05} & \textbf{1.02} & \textbf{8.09} & 77.88  & \textbf{1.73} & \textbf{1.94} & 8.86  \\
    \bottomrule
    \end{tabular}%
}
  \label{tab:ablation}%
\end{table*}%

\subsection{Ablation Study}
\label{exp:ablation}
To evaluate each component in MaC-Cal, we conduct incremental ablation studies on CIFAR-10 and CIFAR-100. Each experiment adds a module to the previous setup to assess its individual contribution, as shown in Tab.\ref{tab:ablation}.

\begin{itemize}
  \item \textbf{Vanilla}: This baseline employs standard one-stage training with a linear classifier and cross-entropy loss, lacking extra structures or perturbations, achieving strong classification but suffering from notable miscalibration.
   
  \item \textbf{Masked Retraining}: Applies random masks to classifier weights in the second stage, slightly improving calibration with stable accuracy.
  
  \item \textbf{Gradient Restriction}: Gradients are restricted to only the active connections defined by the binary mask. This design significantly improves calibration performance but causes slight accuracy degradation.
  
  \item \textbf{Classifier Restructure}: Replaces the linear classifier with a lightweight two-layer bottleneck, maintaining low ECE and AECE without accuracy loss, enhancing fitting under sparse perturbation.
  
  \item \textbf{Static Adaptive Sparsity}: An adaptive sparsity control mechanism is introduced, based on accuracy-confidence discrepancy. This reduces the need for manual hyperparameter tuning and further improves calibration, but with mild training instability and slight accuracy drop.
  
  \item \textbf{Decaying Adaptive Sparsity}: Adds exponentially decaying threshold $\eta_t$ to adaptive sparsity control for flexible early exploration and stable convergence, with top calibration and comparable accuracy.
\end{itemize}

The components of MaC-Cal form a synergistic framework from architecture to training: weight masking regularizes learning, gradient restriction stabilizes optimization, structure redesign enhances expressiveness, and adaptive sparsity dynamically improves calibration. Each module aids calibration individually, while their integration significantly reduces ECE and AECE with minimal loss in accuracy, confirming the method’s effectiveness in both classification and calibration.

\begin{figure}[t]
  \centering
    \centering
    \includegraphics[width=0.7\linewidth]{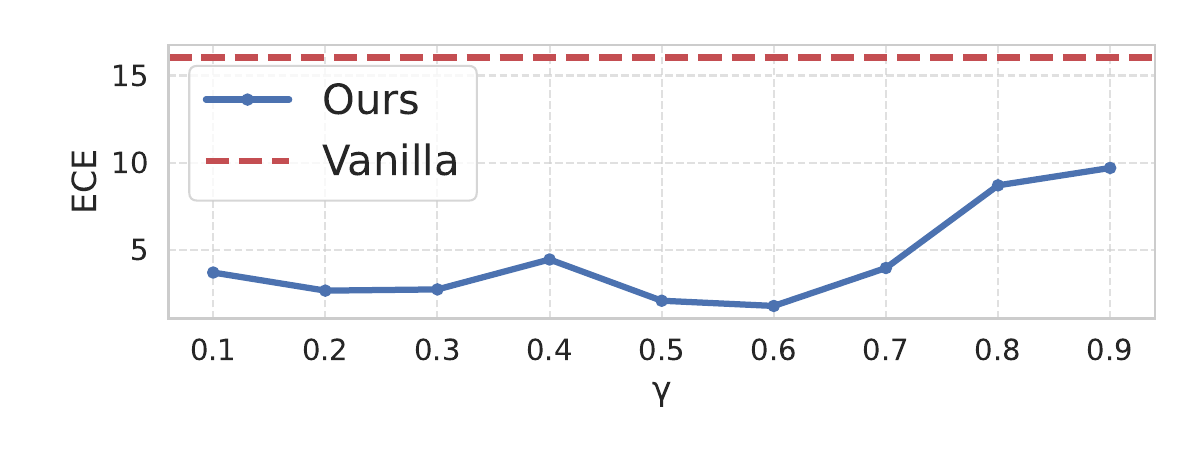}
    \caption{ECE under different values of $\gamma$ on CIFAR-100 using DenseNet-121.}
    \label{fig:gamma}
\end{figure}

\begin{figure}[t]
    \centering
    \includegraphics[width=0.7\linewidth]{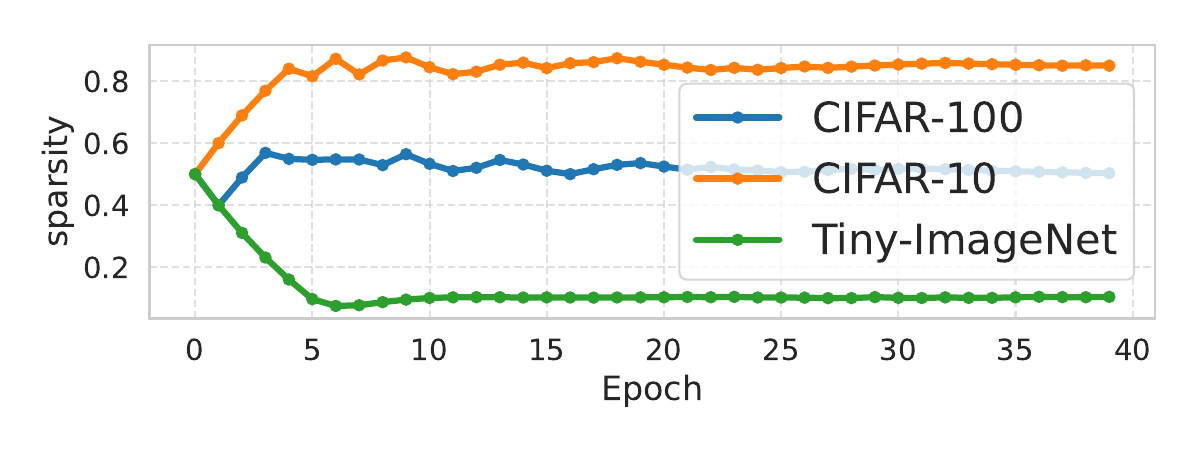}
    \caption{Sparsity variation across training epochs with ResNet-50.}
    \label{fig:epoch}
\end{figure}

% \begin{figure}[t]
%   \centering
%   \subfloat[\label{fig:gamma}]
%   {\includegraphics[width=0.49\textwidth]{fig_fig_gamma.pdf}}
%   \hfill
%   \subfloat[\label{fig:epoch}]
%   {\includegraphics[width=0.49\textwidth]{fig_fig_epoch_sparsity.pdf}}
%   \caption{Parameter Sensitivity. (a) ECE under different values of $\gamma$ on CIFAR-100 using DenseNet-121. (b) Sparsity variation across training epochs with ResNet-50.}
% \end{figure}

\subsection{Parameter Sensitivity}
Manual tuning of fixed sparsity levels is costly and sensitive to model and data variations, limiting generalization and practicality. To overcome this, we propose an adaptive sparsity mechanism with a hyperparameter $\gamma$, allowing sparsity to adjust dynamically during training based on the model’s internal state, thereby eliminating reliance on fixed sparsity configurations.

As shown in Fig.\ref{fig:gamma}, our method improves calibration across a wide range of $\gamma$, indicating low sensitivity and strong robustness. This stability is crucial for achieving consistent performance without extensive tuning. Fig.\ref{fig:epoch} further shows that sparsity adjusts rapidly in early training and stabilizes later, reflecting adaptation to data and architecture. This dynamic process prevents under- or over-regularization seen in fixed strategies.
In summary, the adaptive sparsity improves calibration, enhances model adaptability, and offers a scalable solution for diverse training conditions.
\section{Conclusion}
\label{sec:conclusion}
% In this paper, we propose a mask-based classifier calibration method, denoted as \textbf{MaC-Cal}, which improves confidence calibration by introducing stochastic sparsity into the classifier. Our approach employs a two-stage training strategy: first jointly training the feature extractor and classifier, then freezing the extractor and retraining the classifier under a masking mechanism. Additionally, an adaptive sparsity strategy dynamically adjusts the mask retention probability based on the deviation between predicted confidence and actual accuracy, mitigating both overconfidence and underconfidence issues. Extensive experiments demonstrate that MaC-Cal achieves superior calibration performance with strong generalization and robustness across various models and datasets.
In this paper, we propose \textbf{MaC-Cal}, a mask-based calibration method that improves confidence estimates by introducing stochastic sparsity into the classifier. MaC-Cal adopts a two-stage strategy: joint training of the feature extractor and classifier, followed by classifier retraining with masking. An adaptive sparsity mechanism adjusts the mask retention probability based on confidence–accuracy deviation, addressing both over- and underconfidence. Experiments show that MaC-Cal delivers superior calibration with strong generalization and robustness across models and datasets.

\bibliographystyle{elsarticle-num}
\bibliography{refs}

\end{document}